\pdfoutput=1
\documentclass{article}

\usepackage{microtype}
\usepackage{graphicx}
\usepackage{subfig}
\usepackage{caption}
\usepackage{booktabs} 
\usepackage{amsfonts}
\usepackage{amsmath}

\usepackage{tikz}
\usetikzlibrary{plotmarks}
\usepackage{pgfplots}
\pgfplotsset{compat=1.14}
\pgfplotsset{every tick label/.append style={font=\footnotesize}}

\pgfplotstableread{
x	y	error   x2	y2	error2    x3	y3	error3
4.497	0.331	0.008   4.497	0.189	0.008   4.497	0.079	0.008
12.717	0.241	0.003  12.717	0.074	0.003  12.717	0.025	0.006
21.223	0.204	0.003  21.223	0.033	0.003  21.223	0.025	0.005
29.769	0.177	0.006  29.769	0.025	0.008  29.769	0.032	0.004
38.337	0.150	0.005  38.337	0.038	0.005  38.337	0.036	0.005
47.025	0.133	0.005  47.025	0.058	0.005  47.025	0.041	0.004
56.331	0.122	0.006  56.331	0.067	0.006  56.331	0.036	0.005
66.232	0.108	0.005  66.232	0.081	0.006  66.232	0.034	0.008
76.391	0.095	0.005  76.391	0.096	0.004  76.391	0.034	0.005
87.054	0.084	0.004  87.054	0.105	0.005  87.054	0.031	0.006
98.523	0.074	0.005  98.523	0.118	0.006  98.523	0.033	0.005
110.432	0.070	0.004 110.432	0.121	0.005 110.432	0.026	0.006
122.902	0.057	0.005 122.902	0.134	0.005 122.902	0.027	0.006
136.715	0.054	0.005 136.715	0.137	0.005 136.715	0.025	0.005
153.905	0.051	0.005 153.905	0.140	0.004 153.905	0.023	0.004
176.065	0.048	0.004 176.065	0.142	0.004 176.065	0.024	0.004
206.022	0.044	0.005 206.022	0.145	0.006 206.022	0.026	0.005
250.760	0.035	0.007 250.760	0.154	0.006 250.760	0.033	0.006
327.961	0.034	0.005 327.961	0.158	0.006 327.961	0.041	0.005
555.266	0.034	0.009 555.266	0.188	0.006 555.266	0.036	0.011
}\imdbog

\pgfplotstableread{
x	y	error	  x2	y2	error2    x3	y3	error3	
4.492	0.058	0.005   4.492	0.071	0.004   4.492	0.057	0.005
12.810	0.072	0.010  12.810	0.105	0.008  12.810	0.071	0.009
21.347	0.066	0.007  21.347	0.103	0.009  21.347	0.065	0.007
29.832	0.061	0.007  29.832	0.102	0.007  29.832	0.062	0.008
38.491	0.069	0.013  38.491	0.103	0.013  38.491	0.070	0.012
47.335	0.059	0.010  47.335	0.101	0.005  47.335	0.061	0.008
56.642	0.065	0.012  56.642	0.093	0.019  56.642	0.066	0.013
66.582	0.062	0.014  66.582	0.084	0.018  66.582	0.064	0.011
76.769	0.066	0.016  76.769	0.081	0.016  76.769	0.066	0.016
87.245	0.067	0.009  87.245	0.081	0.020  87.245	0.067	0.008
98.629	0.068	0.014  98.629	0.084	0.018  98.629	0.069	0.014
110.446	0.067	0.014 110.446	0.082	0.019 110.446	0.068	0.015
122.818	0.073	0.017 122.818	0.077	0.019 122.818	0.070	0.017
136.778	0.070	0.014 136.778	0.077	0.018 136.778	0.068	0.013
153.580	0.081	0.015 153.580	0.080	0.018 153.580	0.077	0.016
175.586	0.085	0.019 175.586	0.082	0.020 175.586	0.080	0.017
205.560	0.091	0.022 205.560	0.083	0.021 205.560	0.083	0.021
250.697	0.101	0.026 250.697	0.088	0.025 250.697	0.089	0.026
328.245	0.109	0.024 328.245	0.095	0.024 328.245	0.092	0.023
554.273	0.129	0.024 554.273	0.111	0.024 554.273	0.098	0.023
}\imdbogrnn

\pgfplotstableread{
x	y	error	  x2	y2	error2   x3	y3	error3	
5.103	0.201	0.024   5.103	0.126	0.019   5.103	0.071	0.014
14.579	0.156	0.026  14.579	0.090	0.028  14.579	0.058	0.025
24.523	0.109	0.029  24.523	0.066	0.028  24.523	0.059	0.015
34.475	0.090	0.031  34.475	0.070	0.011  34.475	0.076	0.004
44.370	0.082	0.030  44.370	0.077	0.008  44.370	0.080	0.008
54.342	0.083	0.018  54.342	0.088	0.010  54.342	0.086	0.011
64.864	0.079	0.024  64.864	0.096	0.014  64.864	0.089	0.015
75.879	0.080	0.025  75.879	0.106	0.007  75.879	0.092	0.009
87.454	0.082	0.025  87.454	0.119	0.005  87.454	0.095	0.010
99.941	0.088	0.026  99.941	0.119	0.009  99.941	0.097	0.014
113.009	0.087	0.021 113.009	0.123	0.007 113.009	0.093	0.016
126.460	0.088	0.021 126.460	0.130	0.008 126.460	0.095	0.015
140.542	0.090	0.022 140.542	0.134	0.009 140.542	0.093	0.019
156.308	0.091	0.018 156.308	0.136	0.007 156.308	0.091	0.019
175.118	0.093	0.018 175.118	0.136	0.011 175.118	0.093	0.021
198.820	0.089	0.013 198.820	0.127	0.010 198.820	0.086	0.020
229.861	0.089	0.016 229.861	0.129	0.005 229.861	0.082	0.028
272.862	0.089	0.019 272.862	0.129	0.012 272.862	0.086	0.032
335.824	0.097	0.021 335.824	0.134	0.006 335.824	0.088	0.039
437.525	0.104	0.016 437.525	0.148	0.010 437.525	0.082	0.035
}\imdbogbert

\pgfplotstableread{
x	y	error	  x2	y2	error2   x3	y3	error3	
1.494	0.148	0.010   1.494	0.083	0.006   1.494	0.113	0.007
3.999	0.082	0.003   3.999	0.042	0.006   3.999	0.044	0.002
6.998	0.069	0.004   6.998	0.077	0.006   6.998	0.023	0.004
9.997	0.062	0.004   9.997	0.099	0.007   9.997	0.016	0.003
13.000	0.057	0.004  13.000	0.114	0.004  13.000	0.018	0.004
16.004	0.053	0.003  16.004	0.128	0.006  16.004	0.024	0.006
18.998	0.049	0.003  18.998	0.138	0.005  18.998	0.028	0.005
21.990	0.044	0.004  21.990	0.149	0.006  21.990	0.032	0.006
24.982	0.042	0.003  24.982	0.156	0.005  24.982	0.034	0.006
27.982	0.041	0.002  27.982	0.163	0.004  27.982	0.036	0.005
31.461	0.041	0.003  31.461	0.166	0.005  31.461	0.036	0.005
35.928	0.038	0.004  35.928	0.174	0.005  35.928	0.038	0.006
40.932	0.037	0.003  40.932	0.180	0.005  40.932	0.037	0.005
46.822	0.033	0.004  46.822	0.187	0.004  46.822	0.038	0.006
54.285	0.031	0.002  54.285	0.191	0.004  54.285	0.036	0.005
63.278	0.032	0.004  63.278	0.194	0.004  63.278	0.032	0.005
75.111	0.030	0.003  75.111	0.196	0.004  75.111	0.027	0.006
92.373	0.028	0.003  92.373	0.200	0.003  92.373	0.026	0.008
121.995	0.026	0.003 121.995	0.198	0.002 121.995	0.024	0.004
224.623	0.026	0.004 224.623	0.199	0.002 224.623	0.040	0.002
}\amazonogdan

\pgfplotstableread{
x	y	error	  x2	y2	error2   x3	y3	error3	
1.505	0.268	0.060   1.505	0.285	0.040   1.505	0.275	0.052
3.995	0.099	0.021   3.995	0.140	0.014   3.995	0.114	0.019
6.997	0.037	0.014   6.997	0.068	0.014   6.997	0.047	0.015
10.001	0.026	0.010  10.001	0.045	0.011  10.001	0.027	0.013
13.002	0.022	0.006  13.002	0.034	0.011  13.002	0.020	0.010
15.999	0.021	0.005  15.999	0.030	0.011  15.999	0.018	0.010
18.992	0.020	0.005  18.992	0.028	0.009  18.992	0.017	0.009
21.994	0.019	0.003  21.994	0.026	0.010  21.994	0.016	0.008
24.988	0.017	0.003  24.988	0.023	0.010  24.988	0.015	0.008
28.096	0.018	0.004  28.096	0.023	0.009  28.096	0.015	0.007
31.587	0.019	0.004  31.587	0.022	0.011  31.587	0.015	0.008
35.934	0.020	0.004  35.934	0.022	0.010  35.934	0.016	0.007
40.932	0.020	0.004  40.932	0.022	0.010  40.932	0.016	0.009
46.885	0.021	0.006  46.885	0.023	0.010  46.885	0.018	0.009
54.343	0.022	0.006  54.343	0.023	0.010  54.343	0.018	0.009
63.294	0.023	0.005  63.294	0.023	0.010  63.294	0.019	0.008
75.066	0.024	0.005  75.066	0.023	0.009  75.066	0.020	0.008
92.502	0.026	0.007  92.502	0.024	0.009  92.502	0.022	0.008
122.126	0.028	0.008 122.126	0.026	0.007 122.126	0.026	0.007
223.892	0.036	0.008 223.892	0.029	0.010 223.892	0.034	0.013
}\amazonogrnn

\pgfplotstableread{
x	y	error	  x2	y2	error2   x3	y3	error3	
1.998	0.350	0.070   1.998	0.358	0.084   1.998	0.340	0.054
4.998	0.305	0.051   4.998	0.310	0.057   4.998	0.302	0.043
8.377	0.257	0.047   8.377	0.255	0.051   8.377	0.262	0.042
11.870	0.220	0.048  11.870	0.210	0.050  11.870	0.228	0.044
15.001	0.198	0.048  15.001	0.183	0.053  15.001	0.205	0.045
18.497	0.186	0.049  18.497	0.170	0.054  18.497	0.190	0.048
22.000	0.178	0.047  22.000	0.161	0.052  22.000	0.179	0.046
24.995	0.176	0.047  24.995	0.158	0.053  24.995	0.174	0.048
28.480	0.174	0.044  28.480	0.156	0.048  28.480	0.169	0.045
32.472	0.170	0.042  32.472	0.152	0.047  32.472	0.162	0.044
36.934	0.158	0.046  36.934	0.140	0.050  36.934	0.147	0.048
41.948	0.151	0.042  41.948	0.133	0.047  41.948	0.137	0.046
47.427	0.144	0.043  47.427	0.126	0.048  47.427	0.127	0.047
54.367	0.137	0.044  54.367	0.118	0.049  54.367	0.115	0.050
62.828	0.129	0.047  62.828	0.110	0.052  62.828	0.104	0.052
73.219	0.117	0.042  73.219	0.096	0.047  73.219	0.088	0.045
87.468	0.108	0.045  87.468	0.087	0.052  87.468	0.082	0.041
108.401	0.103	0.042 108.401	0.081	0.049 108.401	0.077	0.031
144.313	0.088	0.041 144.313	0.072	0.037 144.313	0.075	0.017
257.554	0.082	0.044 257.554	0.072	0.029 257.554	0.085	0.013
}\amazonogbert

\pgfplotstableread{
x	y	error   y2	error2   y3	error3   y4	error4
1	0.034	0.011   0.032	0.011   0.034	0.011   0.030	0.009
2	0.029	0.009   0.037	0.009   0.039	0.006   0.041	0.008
3	0.031	0.009   0.036	0.007   0.034	0.008   0.034	0.004
4	0.018	0.007   0.015	0.006   0.018	0.005   0.016	0.008
5	0.047	0.008   0.044	0.008   0.041	0.007   0.041	0.005
6	0.044	0.003   0.033	0.003   0.028	0.001   0.034	0.002
7	0.054	0.010   0.041	0.009   0.025	0.013   0.042	0.008
8	0.068	0.010   0.058	0.008   0.048	0.008   0.060	0.007
9	0.055	0.009   0.042	0.011   0.032	0.011   0.044	0.010
10	0.046	0.003  0.031	0.003  0.019	0.004  0.034	0.005
11	0.046	0.005  0.033	0.003  0.022	0.006  0.032	0.004
12	0.050	0.012  0.041	0.011  0.038	0.012  0.038	0.011
13	0.040	0.006  0.032	0.007  0.032	0.008  0.029	0.006
14	0.032	0.008  0.021	0.008  0.024	0.008  0.020	0.005
15	0.026	0.005  0.020	0.004  0.025	0.003  0.019	0.003
16	0.029	0.003  0.020	0.004  0.028	0.002  0.018	0.006
17	0.023	0.007  0.023	0.007  0.022	0.008  0.025	0.008
18	0.023	0.008  0.031	0.008  0.023	0.008  0.040	0.007
19	0.027	0.003  0.038	0.001  0.025	0.002  0.045	0.002
20	0.022	0.006  0.030	0.006  0.021	0.007  0.040	0.007
21	0.019	0.006  0.022	0.006  0.020	0.006  0.031	0.003
22	0.023	0.004  0.027	0.004  0.022	0.005  0.049	0.005
23	0.024	0.003  0.033	0.005  0.020	0.002  0.044	0.002
24	0.020	0.004  0.027	0.005  0.016	0.004  0.054	0.006
25	0.031	0.006  0.039	0.007  0.027	0.008  0.049	0.009
26	0.026	0.008  0.030	0.009  0.026	0.009  0.053	0.004
27	0.016	0.004  0.018	0.005  0.017	0.004  0.044	0.009
28	0.023	0.004  0.024	0.006  0.024	0.006  0.050	0.006
29	0.028	0.007  0.026	0.007  0.027	0.007  0.060	0.005
30	0.021	0.006  0.015	0.005  0.007	0.003  0.067	0.004
31	0.024	0.004  0.016	0.004  0.008	0.002  0.073	0.003
32	0.031	0.005  0.023	0.004  0.013	0.005  0.081	0.004
33	0.020	0.005  0.014	0.005  0.006	0.002  0.075	0.004
34	0.026	0.005  0.019	0.005  0.005	0.002  0.079	0.005
35	0.029	0.005  0.022	0.005  0.005	0.002  0.083	0.006
}\csgoog

\pgfplotstableread{
x	y1	y2	y3	y4
0.000	1.500	2.310	1.330	1.659
1.000	1.352	2.171	1.405	2.101
2.000	1.226	2.042	1.473	2.398
3.000	1.119	1.924	1.532	2.596
4.000	1.027	1.814	1.585	2.729
5.000	0.949	1.713	1.632	2.819
6.000	0.883	1.620	1.674	2.878
7.000	0.826	1.534	1.711	2.918
8.000	0.778	1.454	1.743	2.945
9.000	0.737	1.381	1.772	2.963
10.000	0.702	1.313	1.798	2.975
11.000	0.672	1.251	1.821	2.984
12.000	0.647	1.193	1.841	2.989
13.000	0.625	1.140	1.859	2.993
14.000	0.606	1.090	1.875	2.995
15.000	0.591	1.045	1.889	2.997
16.000	0.577	1.003	1.902	2.998
17.000	0.566	0.964	1.913	2.999
18.000	0.556	0.929	1.923	2.999
19.000	0.548	0.896	1.931	2.999
20.000	0.541	0.865	1.939	3.000
21.000	0.535	0.837	1.946	3.000
22.000	0.530	0.811	1.952	3.000
23.000	0.525	0.787	1.958	3.000
24.000	0.521	0.765	1.962	3.000
}\expExamples

\pgfplotstableread{
p_t0	p_p0	cp_t0	cp_p0	p_t1	p_p1	cp_t1	cp_p1	p_t2	p_p2	cp_t2	cp_p2	p_t3	p_p3	cp_t3	cp_p3	p_t4	p_p4	cp_t4	cp_p4	p_t5	p_p5	cp_t5	cp_p5	p_t6	p_p6	cp_t6	cp_p6	p_t7	p_p7	cp_t7	cp_p7
0.319	0.274	0.319	0.274	0.099	0.185	0.099	0.135	0.071	0.125	0.071	0.088	0.028	0.061	0.028	0.057	0.000	0.015	0.000	0.016	0.000	0.003	0.000	0.005	0.000	0.003	0.000	0.002	0.000	0.003	0.000	0.000
0.241	0.285	0.241	0.285	0.220	0.245	0.220	0.196	0.142	0.198	0.142	0.156	0.085	0.141	0.085	0.136	0.099	0.071	0.099	0.072	0.014	0.019	0.014	0.025	0.000	0.006	0.000	0.005	0.000	0.006	0.000	0.001
0.397	0.316	0.397	0.315	0.241	0.317	0.241	0.277	0.220	0.270	0.220	0.231	0.234	0.220	0.234	0.215	0.177	0.160	0.177	0.161	0.121	0.085	0.121	0.100	0.007	0.016	0.007	0.014	0.000	0.014	0.000	0.003
0.379	0.410	0.379	0.410	0.293	0.387	0.293	0.360	0.343	0.355	0.343	0.327	0.343	0.322	0.343	0.318	0.300	0.277	0.300	0.277	0.236	0.217	0.236	0.235	0.114	0.123	0.114	0.116	0.000	0.044	0.000	0.016
0.454	0.463	0.454	0.463	0.539	0.463	0.539	0.454	0.362	0.449	0.362	0.439	0.433	0.438	0.433	0.436	0.461	0.422	0.461	0.422	0.482	0.400	0.482	0.408	0.447	0.370	0.447	0.365	0.177	0.228	0.177	0.178
0.546	0.537	0.546	0.537	0.461	0.537	0.461	0.546	0.638	0.551	0.638	0.561	0.567	0.562	0.567	0.564	0.539	0.578	0.539	0.578	0.518	0.600	0.518	0.592	0.553	0.630	0.553	0.635	0.823	0.772	0.823	0.822
0.621	0.590	0.621	0.590	0.707	0.613	0.707	0.640	0.657	0.645	0.657	0.673	0.657	0.678	0.657	0.682	0.700	0.723	0.700	0.723	0.764	0.783	0.764	0.765	0.886	0.877	0.886	0.884	1.000	0.956	1.000	0.984
0.603	0.684	0.603	0.685	0.759	0.683	0.759	0.723	0.780	0.730	0.780	0.769	0.766	0.780	0.766	0.785	0.823	0.840	0.823	0.839	0.879	0.915	0.879	0.900	0.993	0.984	0.993	0.986	1.000	0.986	1.000	0.997
0.759	0.715	0.759	0.715	0.780	0.755	0.780	0.804	0.858	0.802	0.858	0.844	0.915	0.859	0.915	0.864	0.901	0.929	0.901	0.928	0.986	0.981	0.986	0.975	1.000	0.994	1.000	0.995	1.000	0.994	1.000	0.999
0.681	0.726	0.681	0.726	0.901	0.815	0.901	0.865	0.929	0.875	0.929	0.912	0.972	0.939	0.972	0.943	1.000	0.985	1.000	0.984	1.000	0.997	1.000	0.995	1.000	0.997	1.000	0.998	1.000	0.997	1.000	1.000
}\csgoReliability

\pgfplotsset{
  errorBars/.style={
    error bars/error bar style={
      thin,
      solid
    },
    error bars/y dir=both,
    error bars/y explicit,
  },
  legend cell align={left}
}

\usepackage{hyperref}

\usepackage[accepted]{icml2020}

\icmltitlerunning{Temporal Probability Calibration}

\begin{document}

\twocolumn[
\icmltitle{Temporal Probability Calibration}




\begin{icmlauthorlist}
\icmlauthor{Tim Leathart}{ai}
\icmlauthor{Maksymilian Polaczuk}{ai}
\end{icmlauthorlist}

\icmlaffiliation{ai}{Sportsflare AI}
\icmlcorrespondingauthor{Tim Leathart}{tim@sportsflare.io}


\vskip 0.3in
]



\printAffiliationsAndNotice{} 

\begin{abstract}
In many applications, accurate class probability estimates are required, but many types of models produce poor quality probability estimates despite achieving acceptable classification accuracy. Even though probability calibration has been a hot topic of research in recent times, the majority of this has investigated non-sequential data. In this paper, we consider calibrating models that produce class probability estimates from sequences of data, focusing on the case where predictions are obtained from incomplete sequences. We show that traditional calibration techniques are not sufficiently expressive for this task, and propose methods that adapt calibration schemes depending on the length of an input sequence. Experimental evaluation shows that the proposed methods are often substantially more effective at calibrating probability estimates from modern sequential architectures for incomplete sequences across a range of application domains.
\end{abstract}

\section{Introduction}\label{sec:introduction}

Sequential data is abundant in the modern world, commonly seen in forms such as natural language~\cite{harper2016movielens,rajpurkar2016squad}, video streams~\cite{Cordts2016Cityscapes} and financial trading patterns~\cite{brown2013dynamic}. Modern approaches to making predictions from these kinds of data typically involves using model architectures such as deep averaging networks~\cite{iyyer2015deep}, recurrent neural networks~\cite{hochreiter1997long,cho2014learning} and more recently, transformers~\cite{vaswani2017attention,devlin2018bert,radford2019language}. 

In many domains, predicting the most likely class label $\hat{y}_i$ for an instance $i$ with features $\mathbf{x}_i$ and label $y_i$ is sufficient for classification tasks. However, it is often the case in real applications that an estimated probability distribution $\hat{\mathbf{p}}_i$ over the labels can improve the quality or usefulness of the system. For example, an automated loan approval system can be used to minimise expected monetary losses to the lender if the probability of defaulting is accurately estimated. In semi-autonomous vehicles, low estimated confidence for the most likely class for some object may indicate that extra caution is required, possibly alerting the driver to intervene. Sometimes, an accurate probability distribution, rather than a hard classification, is required to have a functioning system at all. For example, if a model is used to predict the winner of a sports game in order to automatically set betting odds, accurate probability estimates are necessary.

\begin{figure}[t]
	\begin{tikzpicture}
		\begin{axis}[
				width=8cm,
	            height=7.5cm,
				ymin=-0.015, ymax=0.365, 
				xmin=3.3, xmax=625,
				ytick={0,0.05,...,0.4}, 
				ytick align=inside, 
				ytick pos=left,
				xtick={0,2,4,8,16,32,64,128,256,512,1024}, 
				xtick align=inside, 
				xtick pos=left,
				xmode=log,
				log basis x=2,
	            grid=major,
	            axis line style=thick,
				yticklabel style={/pgf/number format/fixed},
				xlabel=Sequence Length,
				ylabel=Expected Calibration Error,
				legend pos=north east,
				legend style={nodes={scale=0.65, transform shape}},
				legend cell align={left}
			]
			\addplot [densely dotted, thick, color=red] table [x=x, y=y, col sep=tab] {\imdbog};
			\addlegendentry{Deep Averaging Network};
			\addplot [solid, thick, color=brown] table [x=x, y=y, col sep=tab] {\imdbogrnn};
			\addlegendentry{Recurrent Neural Network};
			\addplot [densely dashed, thick, color=blue] table [x=x, y=y, col sep=tab] {\imdbogbert};
			\addlegendentry{Transformer};
		\end{axis}
	\end{tikzpicture}
	\caption{Test expected calibration error (ECE) of common sequential architectures for sequences of different lengths from the Large Movie Review dataset. As the sequence length increases, the level of calibration for each model changes substantially.\label{fig:intro_ece}}
\end{figure}
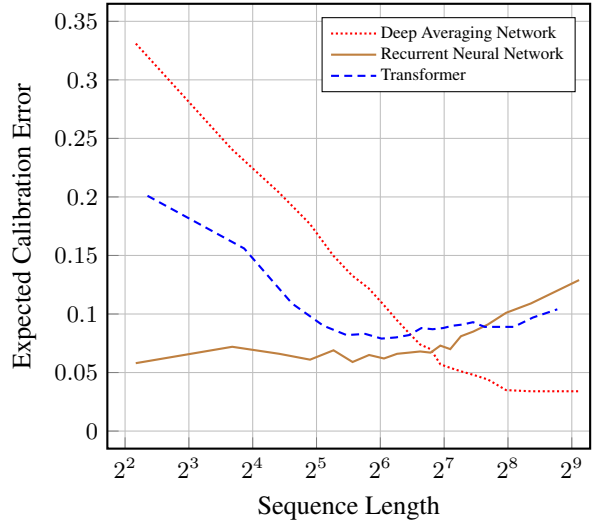

Even though most modern machine learning algorithms natively produce an estimated probability distribution over the class labels for a given instance, it is not always the case that they closely reflect the true probabilities of each class~\cite{niculescu2005predicting,guo2017calibration,leathart2019calibration,kumar2019calibration}. Models for which this is true are said to be poorly \textit{calibrated}. Probability calibration is an additional step one can apply when training a model $f$, where its class probability estimates (or logits) are used as inputs for another model $\pi$ that scales them appropriately to better match the true probabilities.

This work considers situations in which we wish to obtain class probability estimates for a sequence at any time during the formation of the sequence. This situation is fairly common, e.g., offering a ``help'' article to a website user while they are typing a description of the issue they are facing, predicting if an investor should buy or sell an option up until the expiration date, or predicting the outcome of a sports game given information about the current state of the game. For these types of problems, the prediction task typically gets easier as the end of the full sequence draws nearer---if the score in a football game is 1-0 with one minute remaining, we should be much more confident in our prediction of the winner than if the score is 1-0 at half-time. Figure~\ref{fig:intro_ece} shows how expected calibration error, a commonly used calibration metric described in Section~\ref{sec:evaluating_calibration}, changes for the Large Movie Review dataset~\cite{imdb} as the sequence length increases for several different models. In this example, deep average networks and transformers have poorer calibration for shorter sequences than longer, and vice-versa for the recurrent network. Intuitively, a global calibration strategy that applies the same calibration to sequences of any length will not be suitable for these models.

 In this paper, we propose several simple strategies to adapt calibration schemes to better handle incomplete sequences, and evaluate them against traditional, global calibration methods. The paper is structured as follows. First, an introduction to probability calibration is provided, where definitions, existing approaches and evaluation methods are discussed. Then, our proposed temporal probability calibration techniques are described, considering both discrete and continuous sequences of fixed or variable length. Experiments are described and their results discussed. Finally, we go over conclusions and future work.

\section{Probability Calibration}
A probabilistic classifier is said to be perfectly calibrated when the probability estimates for each example exactly match the true class probabilities of the example. For instance, for those examples that are assigned a confidence of $75\%$ by a perfectly calibrated classifier, $75\%$ of them should actually be classified correctly. 

More formally, for a probabilistic classifier $f$ for an $M$-class classification task and predicted probability distribution~$\mathbf{\hat{p}} = [\hat{p}_1, \dots, \hat{p}_M]$, the proportions of classes for all possible instances that would be assigned the prediction~$\mathbf{\hat{p}}$ by~$f$ are equal to~$\mathbf{\hat{p}}$~\cite{kull2019beyond}:
\begin{align}
	\mathbb{P}(Y = m~|~f(\mathbf{X}) = \mathbf{\hat{p}}) = \hat{p}_m \quad \text{for }m \in \{1 \dots M\}.\label{eq:perfect_calibration}
\end{align}

\subsection{Evaluating Calibration\label{sec:evaluating_calibration}}
Without an infinite number of samples, the condition in~(\ref{eq:perfect_calibration}) is not possible to achieve. However, there exist several proxy metrics that aim to emulate this intuition. Common metrics for evaluating the quality of probability estimates include negative log likelihood (NLL)
\begin{align}
	\text{NLL} = - \frac{1}{n} \sum_{i=1}^n \mathbf{y}_i \log \mathbf{\hat{p}}_i
\end{align}
and the Brier score~\cite{brier1950verification}, also known as mean squared error (MSE)
\begin{align}
	\text{MSE} = \frac{1}{n} \sum_{i=1}^n (\mathbf{y}_i - \mathbf{\hat{p}}_i)^2.
\end{align}
Technically speaking, neither of these metrics directly measure calibration, as for each example $i$, the estimated probability $\mathbf{\hat{p}}_i$ is compared to its label $\mathbf{y}_i$ rather than the true probability distribution over the label space. However, they are good proxy metrics, and possess convenient properties for machine learning such as being differentiable, applicable to individual examples, and easy to compute.

Of course, obtaining a true probability distribution over the label space for a single example is usually not possible in practice, as typically, only labels are supplied. Expected calibration error (ECE)~\cite{naeini2015obtaining} is a binning-based approach that attempts to approximate this distribution empirically. The probability space $[0, 1]$ is split into $K$ bins, and the test examples are grouped into the bins based on their estimated probabilities. The accuracy and average confidence of each bin $B_k$ are computed as
\begin{align}
	\text{acc}(B_k) &= \frac{1}{n_k} \sum_i^{n_k} \mathbb{I}(\hat{y}_i = y_i) \\
	\text{conf}(B_k) &= \frac{1}{n_k} \sum_i^{n_k} \hat{p}_i.
\end{align}
ECE is then defined as
\begin{align}
	\text{ECE} = \sum_{k=1}^K \frac{n_k}{n} \big| \text{acc}(B_k) - \text{conf}(B_k) \big|.
\end{align}
For a perfectly calibrated model, the accuracy and confidence of each bin should be equal. Even though ECE measures (average) calibration directly, it is not without problems---the choice for number of bins is arbitrary, probabilities for individual examples are discarded in favour of aggregated bins, and its applicability to multiclass problems is debated~\cite{nixon2019measuring,leathart2019tree}. Strategies such as classwise-ECE~\cite{kull2019beyond} have been proposed to better handle the multiclass case.

Accuracy and confidence are often compared visually in reliability diagrams, where they are plotted against each other~\cite{degroot1983comparison}. In reliability diagrams, perfect calibration is shown by a straight diagonal line. Regions where the curve sits above the diagonal represent underconfidence, and regions where the curve sits under the diagonal represent overconfidence.

\subsection{Parametric Calibration Methods}
One of the most well-known approaches to probability calibration is Platt scaling~\cite{platt1999probabilistic}, in which a univariate logistic regression model with parameters $(\alpha, \beta)$ is learned to minimise NLL between a binary model's outputs $f(\mathbf{x}_i)$ and the labels $y_i$. Calibrated probabilities $\hat{p}_i$ for an instance $i$ can be obtained by
\begin{align}
	\hat{p}_i = \pi(f(\mathbf{x}_i); \alpha, \beta) = \sigma\left(\alpha f(\mathbf{x}_i) + \beta\right)
\end{align}
where $\sigma$ is the sigmoid function. Platt scaling can be applied to multiclass problems using the widely known one-vs-rest method~\cite{zadrozny2002transforming}. Note that $f(\mathbf{x})$, the input to the calibration model, should be a logit rather than a probability. This is because logistic regression assumes a linear relationship between the inputs $f(\mathbf{x})$ and the output logits. 

\citet{guo2017calibration} proposed several variants of Platt scaling---vector scaling, matrix scaling and temperature scaling---that can be applied to multiclass problems. Matrix and vector scaling are identical to multinomial logistic regression where a weights matrix $\mathbf{W}$ and bias vector $\mathbf{b}$ is learned
\begin{align}
	\mathbf{\hat{p}}_i = \pi(f(\mathbf{x}_i); \mathbf{W}, \mathbf{b}) = \sigma \left(\mathbf{W} f(\mathbf{x}_i) + \mathbf{b}\right).
\end{align}
The difference between them is that in vector scaling,~$\mathbf{W}$ is restricted to be a diagonal matrix, while matrix scaling imposes no constraints on~$\mathbf{W}$. Vector scaling can be seen as applying Platt scaling in a one-vs-rest strategy, but with jointly optimised weights. Finally, temperature scaling is the most simple strategy which learns a single parameter $\tau$ (referred to as the temperature) that scales the logits for all classes in the same way:
\begin{align}
	\mathbf{\hat{p}}_i = \pi(f(\mathbf{x}_i); \tau) = \sigma \left(\frac{f(\mathbf{x}_i)}{\tau}\right)
\end{align}
An interesting property of temperature scaling is that because there is no bias term applied, it simply makes probabilistic predictions more or less extreme without changing their classification, and hence does not affect the classification accuracy~\cite{guo2017calibration}.

Beta scaling~\cite{kull2017beyond} and Dirichlet scaling~\cite{kull2019beyond} are, in a practical sense, very similar to Platt scaling and matrix scaling respectively. They model beta and Dirichlet distributions on probabilities $f(\mathbf{x}) \in [0, 1]$, but convert them to logits and log-probabilities respectively and use logistic regression to fit them. A regularisation scheme that penalises off-diagonal and bias parameters was also proposed which lead to strong results for both Dirichlet scaling and matrix scaling~\cite{kull2019beyond}.

\subsection{Nonparametric Calibration Methods}
Histogram binning~\cite{zadrozny2001obtaining} is a simple nonparametric approach to probability calibration. In histogram binning, the model's output space is split into $K$ bins, typically by equal-width or equal-frequency strategies. A calibrated probability per bin is assigned such that the MSE for each bin is minimised, which turns out to be equal to the percentage of positive examples in each bin respectively. The calibrated probability estimate for a test example is given by the assigned value for the bin that it lands in.~\citet{naeini2015obtaining} proposed an extension of histogram binning called Bayesian binning into quantiles, which performs Bayesian model averaging over all possible equal-frequency binning schemes. 

Isotonic regression~\cite{zadrozny2002transforming} learns a piecewise constant function that minimises MSE between the uncalibrated probabilities and labels, with the constraint that it must be monotonically increasing. This can be seen as an extension of histogram binning in that it optimises the bin widths and predictions jointly.~\citet{naeini2016binary} proposed an extension of this where monotonicity is not constrained, but encouraged through regularisation. The removal of the monotonicity constraint leads to overfitting if training is performed to completion; to combat this, the collection of so called near-isotonic regression models produced at each step of training are used in an ensemble, with their predictions weighted by the Bayesian information criterion~\cite{schwarz1978estimating}.

\section{Temporal Probability Calibration}
Temporal probability calibration is motivated by the idea that a model ought to be more or less confident about its predictions at different stages of completion of a sequence. \citet{leathart2017probability} showed that overall calibration can be improved by applying different calibration models in different regions of the input space; this work takes a similar approach in the temporal dimension. 

\subsection{Discrete Fixed-Length Sequences\label{sec:discrete_calibration}}
For classification problems where predictions are made at discrete timesteps from $t \in \{1, \dots, T\}$, it is simple and highly effective to produce a series of $T$ calibration models each parametrised by $\theta_t$, $\pi_1(f; \theta_1), \dots, \pi_T (f; \theta_T)$, corresponding to sequences of each possible length. Calibration of a model output $f(\mathbf{X})$ can be performed by
\begin{align}
	\pi(f(\mathbf{X}), t; \theta) = \pi_t(f(\mathbf{X}); \theta_t)
\end{align}
These calibration models $\pi_1, \dots, \pi_T$ are straightforward to fit by optimising the NLL for a held-out calibration set:
\begin{align}
	\text{arg}\min_{\theta_t}~\text{NLL}\big(\mathbf{y}_t, \pi_{t}(f(\mathbf{X}_t); \theta_t)\big) \quad t \in \{1, \dots, T\}
\end{align}
where $\mathbf{X}_t$ and $\mathbf{y}_t$ are the sequential features and labels respectively of a sample of examples from the calibration set of length $t$.

\subsection{Variable-Length and Continuous Sequences}
For classification problems where predictions are made at any real-valued or integer time $t \in (0, T]$ with possibly infinite $T$, a suitable function  must be chosen to continuously evolve the parameters $\theta$ of a calibration model $\pi : \mathbb{R} \rightarrow [0, 1]$. Throughout this section, we consider temperature scaling as the main calibration method for simplicity of notation; however, the ideas presented are quite general and could, in theory, be applied to other (parametric) calibration methods. 

\begin{figure}
	\centering
	\begin{tikzpicture}
		\begin{axis}[
			width=8.5cm,
	        height=6.5cm,
			ymin=0.35, ymax=3.15, 
			ytick={0.5,1,...,3.5},
			xmin=-1, xmax=25,
			xtick={0,3,...,25}, 
			xticklabels={,,},
			ylabel=Inverse Temperature,
			xlabel=Sequence Length,
	        grid=major,
	        axis line style=thick,
			yticklabel style={/pgf/number format/fixed},
		]
			\addplot [solid, thick, color=black] table [x=x, y=y1, col sep=tab] {\expExamples};
			\addplot [solid, thick, color=black] table [x=x, y=y2, col sep=tab] {\expExamples};
			\addplot [solid, thick, color=black] table [x=x, y=y3, col sep=tab] {\expExamples};
			\addplot [solid, thick, color=black] table [x=x, y=y4, col sep=tab] {\expExamples};
		\end{axis}
	\end{tikzpicture}	
	\caption{Examples of possible forms of (\ref{eq:exp_temperature}).\label{fig:exponential_examples}}
\end{figure}
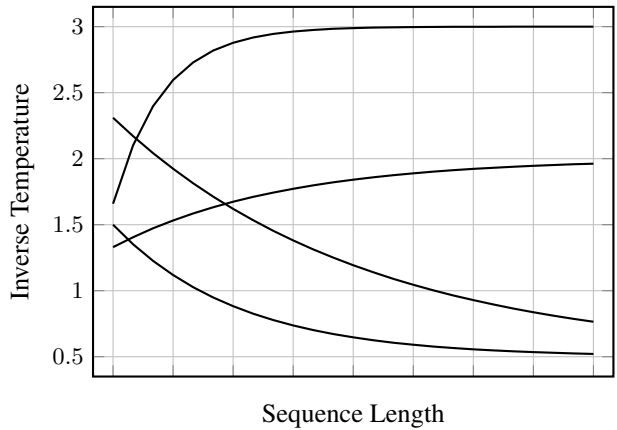

A suitable type of function for modelling an adaptive temperature used in calibration for many problems is a saturating function, such as an exponentially decaying function:
\begin{align}
	g(t; \alpha, \beta, \gamma, s) = \gamma - \alpha e^{(-\beta t - s)} \label{eq:exp_temperature}
\end{align}
which can then be applied in a calibration function like so (parameters of $g$ have been written as $\theta$):
\begin{align}
	\pi(f(\mathbf{X}), t; g, \theta) = \sigma \left( g(t; \theta) f(\mathbf{X})\right).\label{eq:exponential_decay}
\end{align}
The coefficient of $f(\mathbf{X})$ in~(\ref{eq:exponential_decay}) can be interpreted as the inverse of $\tau$ in temperature scaling. A function of this form is flexible enough to allow many different temporal calibration schemes, e.g., continuously decaying toward an upper or lower bound from any starting position.\footnote{This assumes that $\beta$ is positive. One can square this term to enforce this property.} Figure~\ref{fig:exponential_examples} shows some examples of possible temperature functions of this form. A convenient feature of applying temperature scaling in a temporal fashion in this way is that, like in the global calibration case, the total accuracy is left unchanged. 

Similarly to the discrete case, the parameters can be fit to minimise NLL on a held-out calibration set:
\begin{align}
	\text{arg}\min_{\theta}~\text{NLL}\big(\mathbf{y}, \pi(f(\mathbf{X}); g, \theta)\big) \label{eq:fit_continuous}
\end{align}
except that in this case, $\mathbf{X}$ and $\mathbf{y}$ must be the sequential features and labels respectively of a held-out set containing sequences that have been artificially truncated to produce a range of sequence lengths. We found that (\ref{eq:fit_continuous}) can be optimised reliably using typical off-the-shelf optimisers when $t$ is normalised to the range $[0, 1]$, based on the maximum length of sequences in the calibration set. 

\subsection{Alternative Temporal Measures\label{sec:alternative_measures}}

In some problems, it may be more appropriate to use a different indicator of sequence completion than the time $t$ directly. Consider a game played between two players, where the goal is to win five rounds, and a draw is not possible in each round. Clearly, there are a maximum of nine rounds. Intuitively, one might think that as the number of rounds passed increases, we should be more confident of the outcome. However, if a game reaches the ninth round, then the players are likely to be of a similar skill level; thus the probabilities of each outcome becomes more difficult to predict accurately.

A superior measure of sequence completion than the round number $t$ for a game like this may be to fit calibration models to subsets of historical games with equal absolute score difference $s$. Sequences from other domains may have ways of representing an estimation of completion other than directly using the time, and it is left to the practitioner to decide the best approach for their specific situation.

\section{Experimental Results}
In this section, our experimental methods are described and results discussed. As explained in the introduction, there are many situations in which accurate probability estimates from incomplete sequences are useful. However, no datasets for any of these specific tasks exist in the public domain. Nevertheless, we artificially create incomplete sequences from natural language datasets, as well as introduce a sequential dataset for esports outcome prediction, and show that temporal calibration works well across these domains. 

\begin{table*}[t]
	\centering
	\caption{Global performance metrics for natural language datasets. Values presented are the means and standard deviations of ten independent runs. Test examples have been randomly truncated to simulate incomplete sequences. \label{tab:nlp_results}}
	\begin{tabular}{crllcll}
		\addlinespace[-\aboverulesep] \cmidrule[\heavyrulewidth]{3-7} & & \multicolumn{2}{c}{Large Movie Review} && \multicolumn{2}{c}{Amazon Fine Food Review} \\ 
		\cmidrule{3-4} \cmidrule{6-7} 
		      &             & $\quad$~~~ NLL & $\quad$~~~ ECE & &  $\quad$~~~ NLL & $\quad$~~~ ECE \\
	\midrule
			  &	No Calibration    	    & 0.646 $\pm$ 0.008 & 0.107 $\pm$ 0.077 & & 0.412 $\pm$ 0.005 & 0.048 $\pm$ 0.027 $\bullet$  \\
		DAN   &	Global Calibration	    & 0.548 $\pm$ 0.005 $\bullet$ & 0.110 $\pm$ 0.048 & & 0.386 $\pm$ 0.032 & 0.152 $\pm$ 0.046   \\
			  &	Temporal	 Calibration    & \textbf{0.497 $\pm$ 0.004} $\bullet$ & \textbf{0.034 $\pm$ 0.012} $\bullet$ & & \textbf{0.340 $\pm$ 0.007} $\bullet$ & \textbf{0.037 $\pm$ 0.023} $\bullet$\\
	\midrule
			  &	No Calibration	        & 0.464 $\pm$ 0.011 & 0.076 $\pm$ 0.018 $\bullet$ 							  & & 0.233 $\pm$ 0.009  & 0.039 $\pm$ 0.055 $\bullet$\\
		GRU   &	Global Calibration	    & 0.458 $\pm$ 0.010 & 0.089 $\pm$ 0.011 									  & & 0.229 $\pm$ 0.008 $\bullet$ & 0.047 $\pm$ 0.061  \\
			  &	Temporal	 Calibration	    & \textbf{0.450 $\pm$ 0.008} $\bullet$ & \textbf{0.072 $\pm$ 0.010} $\bullet$ & & \textbf{0.226 $\pm$ 0.008} $\bullet$ & \textbf{0.038 $\pm$ 0.058} $\bullet$ \\
	\midrule
			  &	No Calibration	        & 0.593 $\pm$ 0.012 &  0.098 $\pm$ 0.029 $\bullet$							  & &  0.533 $\pm$ 0.045 & 0.172 $\pm$ 0.068    \\
		BERT  &	Global Calibration	    & 0.592 $\pm$ 0.011 &  0.114 $\pm$ 0.024	  								  & &  0.534 $\pm$ 0.047 & \textbf{0.157 $\pm$ 0.075} $\bullet$   \\
			  &	Temporal	 Calibration	    & \textbf{0.577 $\pm$ 0.009} $\bullet$ & \textbf{0.085 $\pm$ 0.011} $\bullet$ & & \textbf{0.521 $\pm$ 0.038} $\bullet$ & 0.162 $\pm$ 0.074 $\bullet$   \\
	\bottomrule		
	\end{tabular}
\end{table*}

For all results, we present the mean of ten runs with ten different random seeds. In our results tables, the method with the best mean score for each metric is bolded. Additionally, we also compute statistical significance by the Friedman test followed by the Nemenyi post-hoc test at $p=0.05$~\cite{demvsar2006statistical}. This is a nonparametric test for multiple classifier comparison that compares average ranks of each method across the ten runs. If the average ranks of two methods are less than a critical difference (a function of number of classifiers, number of runs and desired $p$-value), then they are considered statistically indistinguishable from each other. In our tables, a bullet ($\bullet$) indicates that a method is in the best-performing group.

Experiments were conducted using the PyTorch neural networks framework~\cite{paszke2019pytorch} and Huggingface's transformers library~\cite{Wolf2019HuggingFacesTS} on a Google Cloud Service instance equipped with 16 vCPUs, 60GB of memory and an NVIDIA P100 GPU.

\subsection{Natural Language Datasets}
We use two natural language datasets: Large Movie Review~\cite{imdb} and Amazon Fine Food Review~\cite{mcauley2013amateurs}. Large Movie Review is a collection of film reviews taken from the Internet Movie Database (IMDB)~\cite{imdb}. The classification task is to predict if reviews speak positively or negatively about the film. The dataset is split into 20,000 train, 5,000 calibration and 25,000 test examples. The average length of samples in this dataset is 233 words. Amazon Fine Food Review is a larger dataset of food reviews taken from Amazon. The original dataset has classes from one to five stars; in this investigation, we combine the positive reviews (four and five stars) and the neutral/negative reviews (one to three stars) to form a binary classification problem for simplicity. There are 450,000 train, 50,000 calibration, and 68,484 test examples, with an average length of 85 words.

\subsubsection{Experimental Setup}
\begin{figure*}[t]
	\centering
	\resizebox{!}{0.292\textwidth}{
		\subfloat[DAN\label{fig:dan_imdb}]{
			\begin{tikzpicture}
				\begin{axis}[
						width=8cm,
			            height=7cm,
						ymin=-0.017, ymax=0.367, 
						xmin=3.3, xmax=625,
						ytick={0,0.05,...,0.4}, 
						ytick align=inside, 
						ytick pos=left,
						xtick={0,2,4,8,16,32,64,128,256,512,1024}, 
						xtick align=inside, 
						xtick pos=left,
						xmode=log,
						log basis x=2,
			            grid=major,
			            axis line style=thick,
						yticklabel style={/pgf/number format/fixed},
						xlabel=Sequence Length,
						ylabel=ECE (Large Movie Review)
					]
					\addplot [densely dotted, thick, color=red, error bars/.cd, y fixed, y dir=both, y explicit, error bar style={solid}] table [x=x, y=y, y error=error, col sep=tab] {\imdbog};
					\addplot [densely dashed, color=brown, error bars/.cd, y fixed, y dir=both, y explicit] table [x=x2, y=y2, y error=error2, col sep=tab] {\imdbog};
					\addplot [solid, color=blue, error bars/.cd, y fixed, y dir=both, y explicit, error bar style={solid}] table [x=x3, y=y3, y error=error3, col sep=tab] {\imdbog};
				\end{axis}
			\end{tikzpicture}
		}
	}
	\resizebox{!}{0.292\textwidth}{
		\subfloat[GRU\label{fig:gru_imdb}]{
			\begin{tikzpicture}
				\begin{axis}[
						width=8cm,
			            height=7cm,
						ymin=0.034, ymax=0.166, 
						xmin=3.3, xmax=625,
						ytick={0.04,0.06,...,0.18}, 
						ytick align=inside, 
						ytick pos=left,
						yticklabel style={/pgf/number format/fixed},
						xtick={0,4,8,16,32,64,128,256,512,1024}, 
						xtick align=inside, 
						xtick pos=left,
						xmode=log,
						axis line style=thick,
						log basis x=2,
			            grid=major,
						xlabel=Sequence Length
					]
					\addplot [densely dotted, thick, color=red, error bars/.cd, y fixed, y dir=both, y explicit, error bar style={solid}] table [x=x, y=y, y error=error, col sep=tab] {\imdbogrnn};
					\addplot [densely dashed, color=brown, error bars/.cd, y fixed, y dir=both, y explicit] table [x=x2, y=y2, y error=error2, col sep=tab] {\imdbogrnn};
					\addplot [solid, color=blue, error bars/.cd, y fixed, y dir=both, y explicit, error bar style={solid}] table [x=x3, y=y3, y error=error3, col sep=tab] {\imdbogrnn};
				\end{axis}
			\end{tikzpicture}
		}
	}
	\resizebox{!}{0.292\textwidth}{
		\subfloat[BERT\label{fig:bert_imdb}]{
			\begin{tikzpicture}
				\begin{axis}[
						width=8cm,
			            height=7cm,
						ymin=0.009, ymax=0.271, 
						xmin=3.3, xmax=625,
						ytick={0.02,0.06,...,0.26}, 
						ytick align=inside, 
						ytick pos=left,
						yticklabel style={/pgf/number format/fixed},
						xtick={0,4,8,16,32,64,128,256,512,1024}, 
						xtick align=inside, 
						xtick pos=left,
						xmode=log,
						axis line style=thick,
						log basis x=2,
			            grid=major,
						xlabel=Sequence Length,
						legend pos=north east,
						legend style={nodes={scale=0.65, transform shape}}
					]
					\addplot [densely dotted, thick, color=red, error bars/.cd, y fixed, y dir=both, y explicit, error bar style={solid}] table [x=x, y=y, y error=error, col sep=tab] {\imdbogbert};
					\addlegendentry{No Calibration};
					\addplot [densely dashed, color=brown, error bars/.cd, y fixed, y dir=both, y explicit] table [x=x2, y=y2, y error=error2, col sep=tab] {\imdbogbert};
					\addlegendentry{Global Calibration};
					\addplot [solid, color=blue, error bars/.cd, y fixed, y dir=both, y explicit, error bar style={solid}] table [x=x3, y=y3, y error=error3, col sep=tab] {\imdbogbert};
					\addlegendentry{Temporal Calibration};
				\end{axis}
			\end{tikzpicture}
		}
	}
	\\
	\resizebox{!}{0.292\textwidth}{
		\subfloat[DAN\label{fig:dan_amazon}]{
			\begin{tikzpicture}
				\begin{axis}[
						width=8cm,
			            height=7cm,
						ymin=-0.01, ymax=0.22, 
						xmin=0.8, xmax=325,
						ytick={0,0.03,...,0.21}, 
						ytick align=inside, 
						ytick pos=left,
						xtick={0,1,2,4,8,16,32,64,128,256}, 
						xtick align=inside, 
						xtick pos=left,
						xmode=log,
						log basis x=2,
			            grid=major,
			            axis line style=thick,
						yticklabel style={/pgf/number format/fixed},
						xlabel=Sequence Length,
						ylabel=ECE (Amazon Fine Food Review),
					]
					\addplot [densely dotted, thick, color=red, error bars/.cd, y fixed, y dir=both, y explicit, error bar style={solid}] table [x=x, y=y, y error=error, col sep=tab] {\amazonogdan };
					\addplot [densely dashed, color=brown, error bars/.cd, y fixed, y dir=both, y explicit] table [x=x2, y=y2, y error=error2, col sep=tab] {\amazonogdan};
					\addplot [solid, color=blue, error bars/.cd, y fixed, y dir=both, y explicit, error bar style={solid}] table [x=x3, y=y3, y error=error3, col sep=tab] {\amazonogdan};
				\end{axis}
			\end{tikzpicture}
		}
	}
	\resizebox{!}{0.292\textwidth}{
		\subfloat[GRU\label{fig:gru_amazon}]{
			\begin{tikzpicture}
				\begin{axis}[
						width=8cm,
			            height=7cm,
						ymin=-0.017, ymax=0.377, 
						xmin=0.8, xmax=325,
						ytick={0,0.045,...,0.36}, 
						ytick align=inside, 
						ytick pos=left,
						xtick={0,1,2,4,8,16,32,64,128,256}, 
						xtick align=inside, 
						xtick pos=left,
						xmode=log,
						log basis x=2,
			            grid=major,
			            axis line style=thick,
						yticklabel style={/pgf/number format/fixed},
						xlabel=Sequence Length,
					]
					\addplot [densely dotted, thick, color=red, error bars/.cd, y fixed, y dir=both, y explicit, error bar style={solid}] table [x=x, y=y, y error=error, col sep=tab] {\amazonogrnn};
					\addplot [densely dashed, color=brown, error bars/.cd, y fixed, y dir=both, y explicit] table [x=x2, y=y2, y error=error2, col sep=tab] {\amazonogrnn};
					\addplot [solid, color=blue, error bars/.cd, y fixed, y dir=both, y explicit, error bar style={solid}] table [x=x3, y=y3, y error=error3, col sep=tab] {\amazonogrnn};
				\end{axis}
			\end{tikzpicture}
		}
	}
	\resizebox{!}{0.292\textwidth}{
		\subfloat[BERT\label{fig:bert_amazon}]{
			\begin{tikzpicture}
				\begin{axis}[
						width=8cm,
			            height=7cm,
						ymin=-0.027, ymax=0.517, 
						xmin=0.8, xmax=325,
						ytick={0,0.07,...,0.5}, 
						ytick align=inside, 
						ytick pos=left,
						xtick={-2,-1,0,1,2,4,8,16,32,64,128,256}, 
						xtick align=inside, 
						xtick pos=left,
						xmode=log,
						log basis x=2,
			            grid=major,
			            axis line style=thick,
						yticklabel style={/pgf/number format/fixed},
						xlabel=Sequence Length,
						legend pos=north east,
						legend style={nodes={scale=0.65, transform shape}}
					]
					\addplot [densely dotted, thick, color=red, error bars/.cd, y fixed, y dir=both, y explicit, error bar style={solid}] table [x=x, y=y, y error=error, col sep=tab] {\amazonogbert};
					\addlegendentry{No Calibration};
					\addplot [densely dashed, color=brown, error bars/.cd, y fixed, y dir=both, y explicit] table [x=x2, y=y2, y error=error2, col sep=tab] {\amazonogbert};
					\addlegendentry{Global Calibration};
					\addplot [solid, color=blue, error bars/.cd, y fixed, y dir=both, y explicit, error bar style={solid}] table [x=x3, y=y3, y error=error3, col sep=tab] {\amazonogbert};
					\addlegendentry{Temporal Calibration};
				\end{axis}
			\end{tikzpicture}
		}
	}
	\caption{ECE for Large Movie Review (top) and Amazon Fine Food Review (bottom) over different sequence lengths. Each point and set of error bars represents the mean and standard deviation ECE respectively over ten independent runs. \label{fig:nlp_ece}}
\end{figure*}
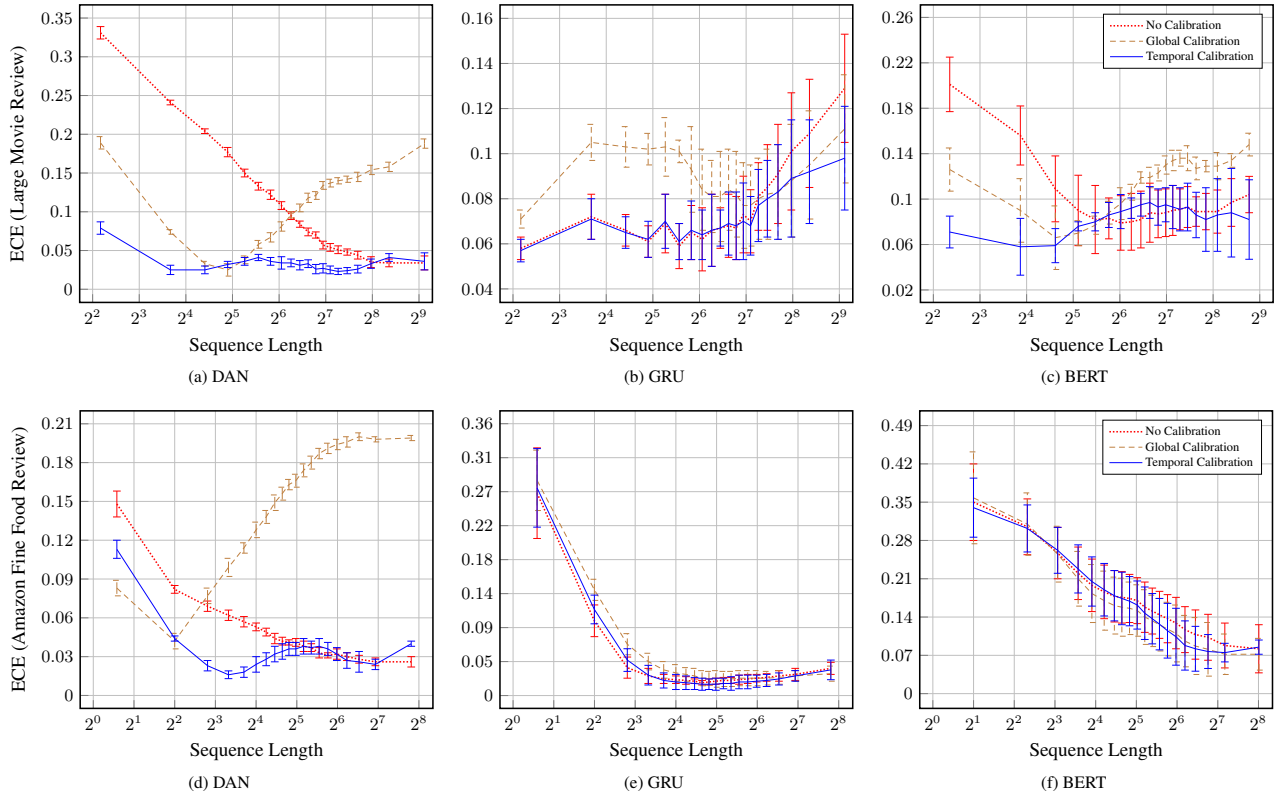

For these datasets, we compare no calibration, global temperature scaling and temporal temperature scaling for three simple natural language processing (NLP) models: a deep averaging network (DAN)~\cite{iyyer2015deep}, recurrent network using gated recurrent units (GRU)~\cite{cho2014learning} and a BERT-based classifier~\citep{devlin2018bert}. Even though transformers are quickly becoming the de-facto models for NLP research, we decided to include DANs and GRUs in this investigation because they have much smaller computational and memory requirements in order to be effective, and are still commonly used in industrial applications. We use 300-dimensional GloVe embeddings to represent the words for DAN and GRU models~\cite{pennington-etal-2014-glove}. 

The DAN had two fully-connected layers after the embedding layer, of 512 and 256 units respectively, before the output layer. The recurrent network contained a GRU followed by one fully-connected layer, each of 256 units, before the output layer. The sum of hidden outputs from the GRU at each time step was passed to the fully-connected. In both of these networks, ReLU activations~\cite{krizhevsky2012learning} and dropout~\cite{srivastava2014dropout} with $p=0.25$ were used between the fully-connected layers. For the BERT-based classifier, we truncate sequences to a maximum length of 512 tokens and use fixed pre-trained weights, only learning the final layer for prediction. Adam~\cite{kingma2014adam} with default settings was used to optimise each network and temporal calibration parameters.

As the sequence length of the examples in these datasets is not uniform, we optimised a continuously decaying exponential function for temporal calibration as in~(\ref{eq:exponential_decay}) in each experiment. The test and calibration sets were constructed by truncating each sequence in the original test set at five random uniformly-sampled points in order to obtain datasets containing a range of sequence completenesses.

\subsubsection{Results and Discussion}
Table~\ref{tab:nlp_results} lists the global NLL and ECE for each calibration method and model for both NLP datasets. Temporal calibration is in the top performing group of results at $p=0.05$ in every comparison.\footnote{In fact, temporal calibration achieved an average rank of exactly one for most comparisons, but the Nemenyi test at $n=10, k=3$ and $p=0.05$ only finds two methods to be statistically indistinguishable if the difference in average ranks is greater than $1.04$. This is why in some cases, pairs of methods with seemingly large NLL and ECE differences are reported as being statistically indistinguishable.} In all but one experiment, the temporal calibration approach achieves the lowest NLL and ECE on the test sets. The effect of temporal calibration is most pronounced for DANs. Interestingly, applying a global calibration scheme often resulted in degraded overall ECE compared to the baseline, despite an improvement in NLL. 

A more enlightening approach to the evaluation of temporal calibration is to visualise calibration for incomplete sequences of different lengths. Figure~\ref{fig:nlp_ece} shows how ECE evolves as the sequence length grows. The altered test sets with truncated examples were binned by sequence length such that each bin contains approximately the same number of samples, and ECE computed for each bin. For all dataset-model pairs, temporal calibration gives the lowest (or tied lowest) ECE for the majority of the time. 

For both datasets, DANs saw marked improvements when using temporal calibration, compared to the baseline as well as global calibration (Figures~\ref{fig:dan_imdb},~\ref{fig:dan_amazon}). The effect of a fine-grained calibration strategy is most obvious for shorter sequences for both datasets. As the sequence length reaches the higher end, temporal calibration matches the performance of the baseline. This may be because a DAN has no notion of sequence length, so it is not able to learn behaviour for specifically dealing with very short sequences, and is often overconfident without the full context available. 

\begin{figure*}
	\centering
		\resizebox{!}{0.272\textwidth}{
		\subfloat[Global Calibration]{
			\begin{tikzpicture}
				\begin{axis}[
					width=8.22cm,
		            height=6.6cm,
					ymin=-0.0035, ymax=0.1035, 
					xmin=0, xmax=37,
					ytick={0,0.02,...,0.14}, 
					ytick align=inside, 
					ytick pos=left,
					xtick={1,6,...,36}, 
					xtick align=inside, 
					xtick pos=left,
		            grid=major,
		            axis line style=thick,
					yticklabel style={/pgf/number format/fixed},
					xlabel=Round Number,
					ylabel=ECE,
					legend pos=north west,
					legend style={nodes={scale=0.65, transform shape}}
				]
					\addplot [densely dotted, thick, color=red, error bars/.cd, y fixed, y dir=both, y explicit, error bar style={solid}] table [x=x, y=y, y error=error, col sep=tab] {\csgoog};
					\addlegendentry{No Calibration};
					\addplot [solid, color=blue, error bars/.cd, y fixed, y dir=both, y explicit] table [x=x, y=y2, y error=error2, col sep=tab] {\csgoog};
					\addlegendentry{Global Calibration};
				\end{axis}
			\end{tikzpicture}
		}
	}
	\resizebox{!}{0.272\textwidth}{
		\subfloat[Temporal Calibration (Round Number)]{
			\begin{tikzpicture}
				\begin{axis}[
					width=8.22cm,
		            height=6.6cm,
					ymin=-0.0035, ymax=0.1035, 
					xmin=0, xmax=37,
					ytick={0,0.02,...,0.14}, 
					ytick align=inside, 
					ytick pos=left,
					yticklabels={,,},
					xtick={1,6,...,36}, 
					xtick align=inside, 
					xtick pos=left,
		            grid=major,
		            axis line style=thick,
					yticklabel style={/pgf/number format/fixed},
					xlabel=Round Number,
					legend pos=north west,
					legend style={nodes={scale=0.65, transform shape}}
				]
					\addplot [densely dotted, thick, color=red, error bars/.cd, y fixed, y dir=both, y explicit, error bar style={solid}] table [x=x, y=y, y error=error, col sep=tab] {\csgoog};
					\addlegendentry{No Calibration};
					\addplot [solid, color=blue, error bars/.cd, y fixed, y dir=both, y explicit] table [x=x, y=y3, y error=error3, col sep=tab] {\csgoog};
					\addlegendentry{Temporal Calibration (Round Number)};
				\end{axis}
			\end{tikzpicture}
		}
	}
	\resizebox{!}{0.272\textwidth}{
		\subfloat[Temporal Calibration (Score Difference)]{
			\begin{tikzpicture}
				\begin{axis}[
					width=8.22cm,
		            height=6.6cm,
					ymin=-0.0035, ymax=0.1035, 
					xmin=0, xmax=37,
					ytick={0,0.02,...,0.14}, 
					ytick align=inside, 
					ytick pos=left,
					yticklabels={,,},
					xtick={1,6,...,36}, 
					xtick align=inside, 
					xtick pos=left,
		            grid=major,
		            axis line style=thick,
					yticklabel style={/pgf/number format/fixed},
					xlabel=Round Number,
					legend pos=north west,
					legend style={nodes={scale=0.65, transform shape}}
				]
					\addplot [densely dotted, thick, color=red, error bars/.cd, y fixed, y dir=both, y explicit, error bar style={solid}] table [x=x, y=y, y error=error, col sep=tab] {\csgoog};
					\addlegendentry{No Calibration};
					\addplot [solid, color=blue, error bars/.cd, y fixed, y dir=both, y explicit] table [x=x, y=y4, y error=error4, col sep=tab] {\csgoog};
					\addlegendentry{Temporal Calibration (Score Difference)};
				\end{axis}
			\end{tikzpicture}
		}
	}
	\caption{ECE of each temporal calibration technique for CS:GO dataset at different round numbers. Each point and set of error bars represents the mean and standard deviation ECE respectively over ten independent runs.\label{fig:csgo_ece}}
\end{figure*}
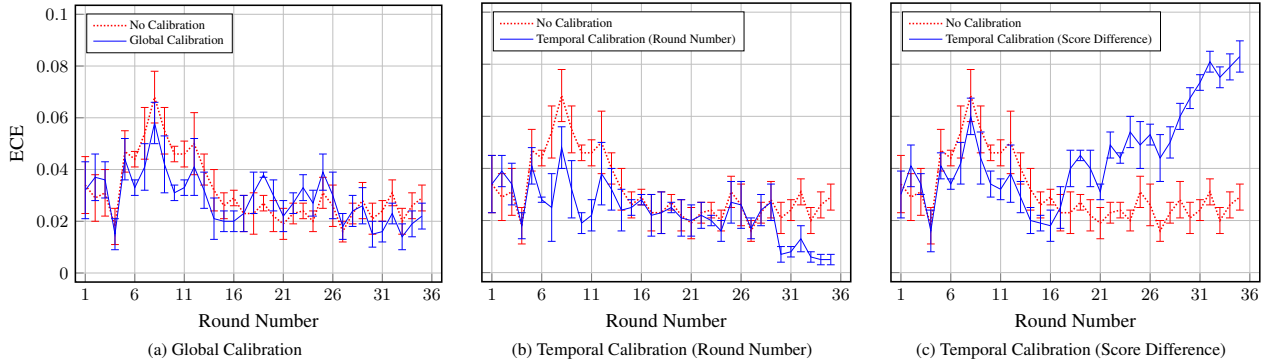

The global calibration scheme baseline had interesting behaviour---usually it matches the performance of temporal scaling for one bin with ECE rising on either side, especially for the smaller dataset, Large Movie Review. It is intuitive that global calibration should behave in this manner, if our assumption that different calibration strategy is required at different sequence lengths is correct, as a global calibration scheme will be unable to adapt to the needs of different regions of temporal space. This result sheds light on how overall ECE can be worse after applying global calibration, despite slight improvements in overall NLL.

In general, calibration appears to be more impactful for smaller datasets when considering GRUs (Figures~\ref{fig:gru_imdb},~\ref{fig:gru_amazon}) and the BERT-based classifier (Figures~\ref{fig:bert_imdb},~\ref{fig:bert_amazon}). We did not see substantial improvements for these models with the larger Amazon Fine Food Review dataset. On the other hand, calibration of these models is greatly improved for Large Movie Reviews. The BERT-based classifier, like the DAN, sees the most improvement for shorter sequences. The calibration curve for BERT (Figure~\ref{fig:bert_imdb}) appears to share a mixture of the behaviours of DAN and GRU (Figures~\ref{fig:dan_imdb},~\ref{fig:gru_imdb}). This may be because the bidirectional self-attention of BERT acts like a weighted average over the input tokens at each timestep similarly to DANs. However, unlike DANs, BERT retains temporal information through its positional encodings. 

The GRU has a stronger sequential assumption than the DAN and BERT built into its architecture. Temporal calibration of GRUs matches the performance of the baseline for shorter sequences, only seeing a reduction in ECE for longer sequences. The performance drop for longer sequences is likely due to the GRU ``forgetting'' about the first tokens as the sequences grow. However, the network has learned to make low-confidence predictions at the beginning of sequences, so there is little to gain from calibration here.

\subsection{CS:GO Round Sequences}
\begin{table}[t]
	\centering
	\small 
	\caption{Global metrics for CS:GO dataset. Values presented are the means and standard deviations of ten independent runs. Test examples have been randomly truncated to simulate incomplete sequences. \label{tab:csgo}}
	\begin{tabular}{cll}
		\toprule
		Calibration Method & $\qquad$ NLL & $\qquad$ ECE \\
		\midrule
		No Calibration & 0.3830 $\pm$ 0.006 & 0.0316 $\pm$ 0.013   \\
		Global Calibration & 0.3815 $\pm$ 0.006 & 0.0290 $\pm$ 0.012 $\bullet$\\
		Temporal (Round) & \textbf{0.3793 $\pm$ 0.006} $\bullet$ & \textbf{0.0235 $\pm$ 0.012} $\bullet$  \\
		Temporal (Score) & 0.4062 $\pm$ 0.004 & 0.0450 $\pm$ 0.018 \\
		\bottomrule
	\end{tabular}
\end{table}

Counter-Strike: Global Offensive (CS:GO) is a first-person shooter multiplayer video game in which two teams of five players play against each other. A game is comprised of two halves of fifteen rounds, where the winning team is the first team to win sixteen rounds. In the event of a draw, a six-round overtime is added to the match until a winner is decided. 

We introduce the first public dataset for CS:GO game winner prediction. It contains sequences of round statistics where each sequence element includes information such as the round winner, number of surviving players and current score. There are 12,362 games in the dataset, which we split into 8,751 training samples, 2,907 calibration samples and 704 test samples. More information about the dataset can be found in the supplementary material.

\subsubsection{Experimental Setup}
\begin{figure*}
	\centering
	\resizebox{!}{0.23\textwidth}{
		\subfloat{
			\begin{tikzpicture}
				\begin{axis}[
					width=7.8cm,
		            height=7.5cm,
					ymin=-0.025, ymax=1.025, 
					xmin=-0.025, xmax=1.025,
					ytick={0,0.2,...,1.2}, 
					ytick pos=left,
					xtick={0,0.2,...,1.2}, 
					xtick pos=left,
					xticklabels={,,},
		            grid=major,
		            axis line style=thick,
					yticklabel style={/pgf/number format/fixed},
					ylabel=Accuracy,
					legend pos=north west,
					legend style={nodes={scale=0.65, transform shape}}
				]
					\addplot [dashed, thin, color=gray, forget plot] coordinates {(-0.5,-0.5) (1.5,1.5)};
					\addplot [densely dotted, thick, color=red, mark=*, mark options={scale=0.3}] table [x=p_p0, y=p_t0, col sep=tab] {\csgoReliability};
					\addlegendentry{No Calibration};
					\addplot [solid, thick, color=blue, mark=*, mark options={scale=0.3}] table [x=cp_p0, y=cp_t0, col sep=tab] {\csgoReliability};
					\addlegendentry{Temporal Calibration};

					\node[] at (axis cs: 0.75,0.04) {Round 2, $\tau=0.948$};
				\end{axis}
			\end{tikzpicture}
		}
	}
	\resizebox{!}{0.23\textwidth}{
		\subfloat{
			\begin{tikzpicture}
				\begin{axis}[
					width=7.8cm,
		            height=7.5cm,
					ymin=-0.025, ymax=1.025, 
					xmin=-0.025, xmax=1.025,
					ytick={0,0.2,...,1.2}, 
					ytick pos=left,
					xtick={0,0.2,...,1.2}, 
					xtick pos=left,
					xticklabels={,,},
					yticklabels={,,},
		            grid=major,
		            axis line style=thick,
					yticklabel style={/pgf/number format/fixed},
				]
					\addplot [dashed, thin, color=gray, forget plot] coordinates {(-0.5,-0.5) (1.5,1.5)};
					\addplot [densely dotted, thick, color=red, mark=*, mark options={scale=0.3}] table [x=p_p1, y=p_t1, col sep=tab] {\csgoReliability};
					\addplot [solid, thick, color=blue, mark=*, mark options={scale=0.3}] table [x=cp_p1, y=cp_t1, col sep=tab] {\csgoReliability};
					\node[] at (axis cs: 0.75,0.04) {Round 6, $\tau=0.799$};
				\end{axis}
			\end{tikzpicture}
		}
	}
	\resizebox{!}{0.23\textwidth}{
		\subfloat{
			\begin{tikzpicture}
				\begin{axis}[
					width=7.8cm,
		            height=7.5cm,
					ymin=-0.025, ymax=1.025, 
					xmin=-0.025, xmax=1.025,
					ytick={0,0.2,...,1.2}, 
					ytick pos=left,
					xtick={0,0.2,...,1.2}, 
					xtick pos=left,
					xticklabels={,,},
					yticklabels={,,},
		            grid=major,
		            axis line style=thick,
					yticklabel style={/pgf/number format/fixed}
				]
					\addplot [dashed, thin, color=gray, forget plot] coordinates {(-0.5,-0.5) (1.5,1.5)};
					\addplot [densely dotted, thick, color=red, mark=*, mark options={scale=0.3}] table [x=p_p2, y=p_t2, col sep=tab] {\csgoReliability};
					\addplot [solid, thick, color=blue, mark=*, mark options={scale=0.3}] table [x=cp_p2, y=cp_t2, col sep=tab] {\csgoReliability};
					\node[] at (axis cs: 0.735,0.04) {Round 10, $\tau=0.838$};
				\end{axis}
			\end{tikzpicture}
		}
	}
	\resizebox{!}{0.23\textwidth}{
		\subfloat{
			\begin{tikzpicture}
				\begin{axis}[
					width=7.8cm,
		            height=7.5cm,
					ymin=-0.025, ymax=1.025, 
					xmin=-0.025, xmax=1.025,
					ytick={0,0.2,...,1.2}, 
					ytick pos=left,
					xtick={0,0.2,...,1.2}, 
					xtick pos=left,
					xticklabels={,,},
					yticklabels={,,},
		            grid=major,
		            axis line style=thick,
					yticklabel style={/pgf/number format/fixed},
				]
					\addplot [dashed, thin, color=gray, forget plot] coordinates {(-0.5,-0.5) (1.5,1.5)};
					\addplot [densely dotted, thick, color=red, mark=*, mark options={scale=0.3}] table [x=p_p3, y=p_t3, col sep=tab] {\csgoReliability};
					\addplot [solid, thick, color=blue, mark=*, mark options={scale=0.3}] table [x=cp_p3, y=cp_t3, col sep=tab] {\csgoReliability};
					\node[] at (axis cs: 0.735,0.04) {Round 14, $\tau=0.989$};
				\end{axis}
			\end{tikzpicture}
		}
	} \\
	\resizebox{!}{0.2545\textwidth}{
		\subfloat{
			\begin{tikzpicture}
				\begin{axis}[
					width=7.8cm,
		            height=7.5cm,
					ymin=-0.025, ymax=1.025, 
					xmin=-0.025, xmax=1.025,
					ytick={0,0.2,...,1.2}, 
					ytick pos=left,
					xtick={0,0.2,...,1.2}, 
					xtick pos=left,
		            grid=major,
		            axis line style=thick,
					yticklabel style={/pgf/number format/fixed},
					xlabel=Confidence,
					ylabel=Accuracy
				]
					\addplot [dashed, thin, color=gray, forget plot] coordinates {(-0.5,-0.5) (1.5,1.5)};
					\addplot [densely dotted, thick, color=red, mark=*, mark options={scale=0.3}] table [x=p_p4, y=p_t4, col sep=tab] {\csgoReliability};
					\addplot [solid, thick, color=blue, mark=*, mark options={scale=0.3}] table [x=cp_p4, y=cp_t4, col sep=tab] {\csgoReliability};
					\node[] at (axis cs: 0.735,0.04) {Round 18, $\tau=1.014$};
				\end{axis}
			\end{tikzpicture}
		}
	}
	\resizebox{!}{0.2535\textwidth}{
		\subfloat{
			\begin{tikzpicture}
				\begin{axis}[
					width=7.8cm,
		            height=7.5cm,
					ymin=-0.025, ymax=1.025, 
					xmin=-0.025, xmax=1.025,
					ytick={0,0.2,...,1.2}, 
					ytick pos=left,
					xtick={0,0.2,...,1.2}, 
					xtick pos=left,
		            grid=major,
		            axis line style=thick,
		            yticklabels={,,},
					xlabel=Confidence,
				]
					\addplot [dashed, thin, color=gray, forget plot] coordinates {(-0.5,-0.5) (1.5,1.5)};
					\addplot [densely dotted, thick, color=red, mark=*, mark options={scale=0.3}] table [x=p_p5, y=p_t5, col sep=tab] {\csgoReliability};
					\addplot [solid, thick, color=blue, mark=*, mark options={scale=0.3}] table [x=cp_p5, y=cp_t5, col sep=tab] {\csgoReliability};
					\node[] at (axis cs: 0.735,0.04) {Round 22, $\tau=1.080$};
				\end{axis}
			\end{tikzpicture}
		}
	}
	\resizebox{!}{0.2535\textwidth}{
		\subfloat{
			\begin{tikzpicture}
				\begin{axis}[
					width=7.8cm,
		            height=7.5cm,
					ymin=-0.025, ymax=1.025, 
					xmin=-0.025, xmax=1.025,
					ytick={0,0.2,...,1.2},  
					ytick pos=left,
					xtick={0,0.2,...,1.2}, 
					xtick pos=left,
		            grid=major,
		            axis line style=thick,
					yticklabels={,,},
					xlabel=Confidence
				]
					\addplot [dashed, thin, color=gray, forget plot] coordinates {(-0.5,-0.5) (1.5,1.5)};
					\addplot [densely dotted, thick, color=red, mark=*, mark options={scale=0.3}] table [x=p_p6, y=p_t6, col sep=tab] {\csgoReliability};
					\addplot [solid, thick, color=blue, mark=*, mark options={scale=0.3}] table [x=cp_p6, y=cp_t6, col sep=tab] {\csgoReliability};
					\node[] at (axis cs: 0.735,0.04) {Round 26, $\tau=0.961$};
				\end{axis}
			\end{tikzpicture}
		}
	}
	\resizebox{!}{0.2535\textwidth}{
		\subfloat{
			\begin{tikzpicture}
				\begin{axis}[
					width=7.8cm,
		            height=7.5cm,
					ymin=-0.025, ymax=1.025, 
					xmin=-0.025, xmax=1.025,
					ytick={0,0.2,...,1.2}, 
					ytick pos=left,
					xtick={0,0.2,...,1.2}, 
					xtick pos=left,
		            grid=major,
		            axis line style=thick,
					yticklabels={,,},
					xlabel=Confidence,
				]
					\addplot [dashed, thin, color=gray, forget plot] coordinates {(-0.5,-0.5) (1.5,1.5)};
					\addplot [densely dotted, thick, color=red, mark=*, mark options={scale=0.3}] table [x=p_p7, y=p_t7, col sep=tab] {\csgoReliability};
					\addplot [solid, thick, color=blue, mark=*, mark options={scale=0.3}] table [x=cp_p7, y=cp_t7, col sep=tab] {\csgoReliability};
					\node[] at (axis cs: 0.735,0.04) {Round 30, $\tau=0.749$};
				\end{axis}
			\end{tikzpicture}
		}
	}
	\caption{Reliability diagrams of round-based temporal calibration and the baseline for the CS:GO dataset at different rounds. Temperature~($\tau$), fitted from the calibration set, is listed for each round number.\label{fig:csgo_reliability}}
\end{figure*}
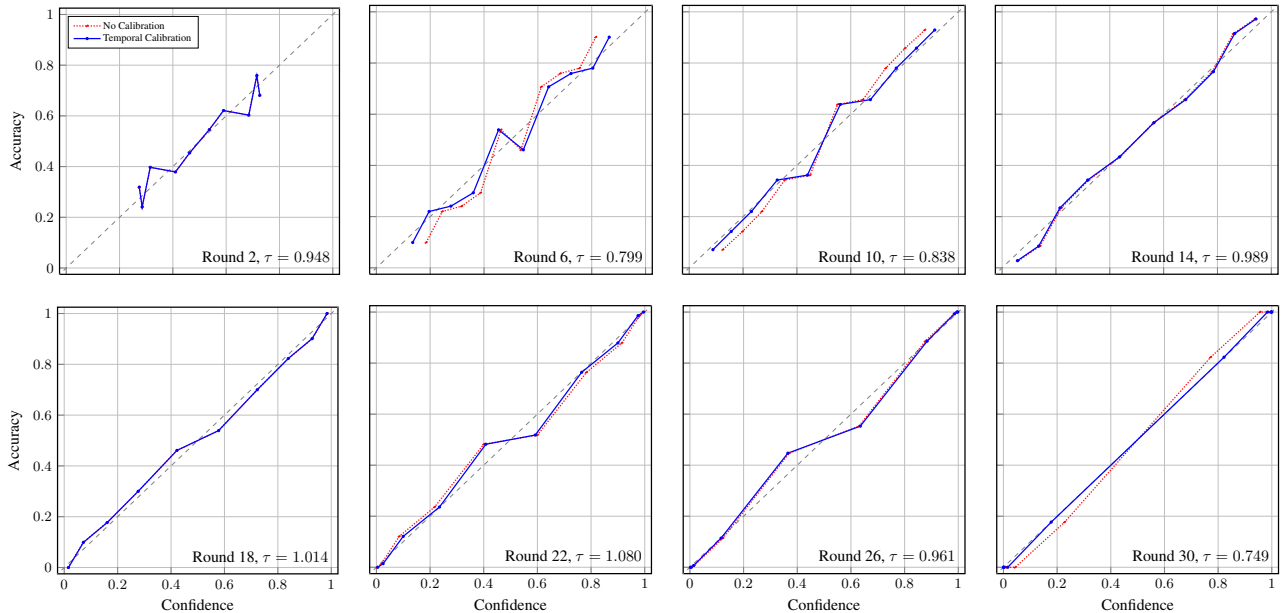
For this dataset, we only use a recurrent neural network with gated recurrent units~\cite{cho2014learning} that uses the raw features as inputs. The GRU has 96 units, followed by two feed-forward fully-connected layers of 96 units each. ReLU activations~\cite{krizhevsky2012learning} are applied between the linear fully-connected layers, as well as dropout~\cite{srivastava2014dropout} with $p=0.25$. 

As before, no calibration and global temperature scaling are used as baselines. These sequences have a maximum length and discrete timesteps, so we use discrete temporal calibration as described in Section~\ref{sec:discrete_calibration}. Temperature scaling is used to calibrate each timestep. We experiment with using the round number and the absolute score difference as measures of time. As with the NLP datasets, we produce an augmented test and calibration set where examples are randomly truncated.

\subsubsection{Results and Discussion}
Table~\ref{tab:csgo} shows global NLL and ECE of each calibration method for the CS:GO dataset. For this data, global temperature scaling shows slight improvements overall compared to the baseline, and temporal scaling based on round number achieves the best results. Interestingly, temporal scaling based on score difference has the worst calibration overall with a degradation compared to the baseline.

Figure~\ref{fig:csgo_ece} shows how ECE evolves with the length of the sequence. While global calibration does achieve slight ECE improvements between rounds five and sixteen, as well as improvements at the end of the game and during overtime, the calibration degrades in-between these areas. Again, it is clear that global calibration strategies are suboptimal. Temporal calibration by round number results in lower ECE than global calibration in these areas, while matching the performance of the baseline during the mid-game rounds. Temporal calibration by score difference, hypothesised in Section~\ref{sec:alternative_measures} to be more appropriate than round number for this type of sequence, turned out to have relatively poor performance, with ECE rising fairly steadily above that of the baseline after round sixteen. This may be due to the score difference not being able to differentiate between, for example, an absolute score difference of two at round two as opposed to at the penultimate round. The truncation strategy employed results in comparatively more short sequences than long ones, which may explain why calibration degrades near the end of the games.

Figure~\ref{fig:csgo_reliability} shows reliability diagrams comparing the baseline to temporal calibration (by round number). Each plot shows the calibration at a different round number, illustrating the progression of the predictions across the length of a whole game. In reliability diagrams, perfect calibration is shown by a perfect diagonal line. We use ten equal-frequency bins in these plots. Even though the calibration of the baseline at each step is reasonably good, visible improvements are made at rounds six, ten, and thirty, while the calibration of the middle rounds stays virtually the same. As equal-frequency binning and temperature scaling are employed for each round number, the accuracy of each bin (on the $y$-axis) does not change, but the confidence (on the $x$-axis) moves closer to the diagonal line after calibration is applied.

\section{Conclusion}
In this paper, we investigated probability calibration for sequential data; specifically looking at how calibration of the classification of a sequence that is being produced changes over time. Two simple methods for calibrating different types of sequences---exponentially-decaying temperature for continuous sequences and timestep-based binning for discrete sequences---were proposed, each showing an improvement in NLL and ECE in their respective areas of application. Especially high performance is obtained when a DAN is used for natural language data, or a relatively low amount of training data is available. 

This paper touches on the calibration of predictions made by transformer-based classifiers, specifically BERT; an interesting avenue of future research would be to perform a thorough investigation on this topic. Some work has been done in this area for neural machine translation~\cite{kumar2019calibration, muller2019does}, but further investigation for standard classification tasks would be a valuable research contribution. 

\section*{Acknowledgements}
The authors thank Christopher Laing and Chris Herrmann for useful discussions. The authors also thank Google for providing cloud credits that were used to run the experiments using Google Cloud Platform.


\bibliography{example_paper}

\begin{thebibliography}{37}
\providecommand{\natexlab}[1]{#1}
\providecommand{\url}[1]{\texttt{#1}}
\expandafter\ifx\csname urlstyle\endcsname\relax
  \providecommand{\doi}[1]{doi: #1}\else
  \providecommand{\doi}{doi: \begingroup \urlstyle{rm}\Url}\fi

\bibitem[Brier(1950)]{brier1950verification}
Brier, G.~W.
\newblock Verification of forecasts expressed in terms of probability.
\newblock \emph{Monthly Weather Review}, 78\penalty0 (1):\penalty0 1--3, 1950.

\bibitem[Brown et~al.(2013)Brown, Pelosi, and Dirska]{brown2013dynamic}
Brown, M.~S., Pelosi, M.~J., and Dirska, H.
\newblock Dynamic-radius species-conserving genetic algorithm for the financial
  forecasting of {D}ow {J}ones index stocks.
\newblock In \emph{Proceedings of the International Workshop on Machine
  Learning and Data Mining in Pattern Recognition}, pp.\  27--41. Springer,
  2013.

\bibitem[Cho et~al.(2014)Cho, van Merrienboer, Gulcehre, Bahdanau, Bougares,
  Schwenk, and Bengio]{cho2014learning}
Cho, K., van Merrienboer, B., Gulcehre, C., Bahdanau, D., Bougares, F.,
  Schwenk, H., and Bengio, Y.
\newblock Learning phrase representations using {RNN} encoder--decoder for
  statistical machine translation.
\newblock In \emph{Proceedings of the Conference on Empirical Methods in
  Natural Language Processing}, pp.\  1724--1734, 2014.

\bibitem[Cordts et~al.(2016)Cordts, Omran, Ramos, Rehfeld, Enzweiler, Benenson,
  Franke, Roth, and Schiele]{Cordts2016Cityscapes}
Cordts, M., Omran, M., Ramos, S., Rehfeld, T., Enzweiler, M., Benenson, R.,
  Franke, U., Roth, S., and Schiele, B.
\newblock The cityscapes dataset for semantic urban scene understanding.
\newblock In \emph{Proceedings of the 29th IEEE Conference on Computer Vision
  and Pattern Recognition}, 2016.

\bibitem[DeGroot \& Fienberg(1983)DeGroot and Fienberg]{degroot1983comparison}
DeGroot, M.~H. and Fienberg, S.~E.
\newblock The comparison and evaluation of forecasters.
\newblock \emph{Journal of the Royal Statistical Society: Series D (The
  Statistician)}, 32\penalty0 (1-2):\penalty0 12--22, 1983.

\bibitem[Dem{\v{s}}ar(2006)]{demvsar2006statistical}
Dem{\v{s}}ar, J.
\newblock Statistical comparisons of classifiers over multiple data sets.
\newblock \emph{Journal of Machine Learning Research}, 7\penalty0
  (Jan):\penalty0 1--30, 2006.

\bibitem[Devlin et~al.(2018)Devlin, Chang, Lee, and Toutanova]{devlin2018bert}
Devlin, J., Chang, M.-W., Lee, K., and Toutanova, K.
\newblock {BERT}: Pre-training of deep bidirectional transformers for language
  understanding.
\newblock \emph{arXiv preprint arXiv:1810.04805}, 2018.

\bibitem[Guo et~al.(2017)Guo, Pleiss, Sun, and Weinberger]{guo2017calibration}
Guo, C., Pleiss, G., Sun, Y., and Weinberger, K.~Q.
\newblock On calibration of modern neural networks.
\newblock In \emph{Proceedings of the 34th International Conference on Machine
  Learning}, pp.\  1321--1330, 2017.

\bibitem[Harper \& Konstan(2016)Harper and Konstan]{harper2016movielens}
Harper, F.~M. and Konstan, J.~A.
\newblock The movielens datasets: History and context.
\newblock \emph{ACM Transactions on Interactive Intelligent Systems},
  5\penalty0 (4):\penalty0 19, 2016.

\bibitem[Hochreiter \& Schmidhuber(1997)Hochreiter and
  Schmidhuber]{hochreiter1997long}
Hochreiter, S. and Schmidhuber, J.
\newblock Long short-term memory.
\newblock \emph{Neural computation}, 9\penalty0 (8):\penalty0 1735--1780, 1997.

\bibitem[Iyyer et~al.(2015)Iyyer, Manjunatha, Boyd-Graber, and
  Daum{\'e}~III]{iyyer2015deep}
Iyyer, M., Manjunatha, V., Boyd-Graber, J., and Daum{\'e}~III, H.
\newblock Deep unordered composition rivals syntactic methods for text
  classification.
\newblock In \emph{Proceedings of the Annual Meeting of the Association for
  Computational Linguistics and the 7th International Joint Conference on
  Natural Language Processing}, pp.\  1681--1691, 2015.

\bibitem[Kingma \& Ba(2015)Kingma and Ba]{kingma2014adam}
Kingma, D.~P. and Ba, J.
\newblock Adam: {A} method for stochastic optimization.
\newblock In \emph{Proceedings of the 3rd International Conference on Learning
  Representations}, 2015.

\bibitem[Krizhevsky et~al.(2012)Krizhevsky, Sutskever, and
  Hinton]{krizhevsky2012learning}
Krizhevsky, A., Sutskever, I., and Hinton, G.~E.
\newblock Image{N}et classification with deep convolutional neural networks.
\newblock In \emph{Proceedings of Advances in Neural Information Processing
  Systems}, pp.\  1097--1105, 2012.

\bibitem[Kull et~al.(2017)Kull, Silva~Filho, Flach, et~al.]{kull2017beyond}
Kull, M., Silva~Filho, T.~M., Flach, P., et~al.
\newblock Beyond sigmoids: How to obtain well-calibrated probabilities from
  binary classifiers with beta calibration.
\newblock \emph{Electronic Journal of Statistics}, 11\penalty0 (2):\penalty0
  5052--5080, 2017.

\bibitem[Kull et~al.(2019)Kull, Nieto, K{\"a}ngsepp, Silva~Filho, Song, and
  Flach]{kull2019beyond}
Kull, M., Nieto, M.~P., K{\"a}ngsepp, M., Silva~Filho, T., Song, H., and Flach,
  P.
\newblock Beyond temperature scaling: Obtaining well-calibrated multi-class
  probabilities with {D}irichlet calibration.
\newblock In \emph{Proceedings of Advances in Neural Information Processing
  Systems}, pp.\  12295--12305, 2019.

\bibitem[Kumar \& Sarawagi(2019)Kumar and Sarawagi]{kumar2019calibration}
Kumar, A. and Sarawagi, S.
\newblock Calibration of encoder decoder models for neural machine translation.
\newblock \emph{arXiv preprint arXiv:1903.00802}, 2019.

\bibitem[Leathart et~al.(2017)Leathart, Frank, Holmes, and
  Pfahringer]{leathart2017probability}
Leathart, T., Frank, E., Holmes, G., and Pfahringer, B.
\newblock Probability calibration trees.
\newblock In \emph{Proceedings of the 9th Asian Conference on Machine
  Learning}, pp.\  145--160. PMLR, 2017.

\bibitem[Leathart et~al.(2019)Leathart, Frank, Pfahringer, and
  Holmes]{leathart2019calibration}
Leathart, T., Frank, E., Pfahringer, B., and Holmes, G.
\newblock On calibration of nested dichotomies.
\newblock In \emph{Proceedings of the 23rd Pacific-Asia Conference on Knowledge
  Discovery and Data Mining}, pp.\  69--80. Springer, 2019.

\bibitem[Leathart(2019)]{leathart2019tree}
Leathart, T.~M.
\newblock \emph{Tree-structured multiclass probability estimators}.
\newblock PhD thesis, University of Waikato, 2019.

\bibitem[Maas et~al.()Maas, Daly, Pham, Huang, Ng, and Potts]{imdb}
Maas, A.~L., Daly, R.~E., Pham, P.~T., Huang, D., Ng, A.~Y., and Potts, C.
\newblock Learning word vectors for sentiment analysis.
\newblock In \emph{Proceedings of the 49th Annual Meeting of the Association
  for Computational Linguistics: Human Language Technologies}, pp.\  142--150.

\bibitem[McAuley \& Leskovec(2013)McAuley and Leskovec]{mcauley2013amateurs}
McAuley, J.~J. and Leskovec, J.
\newblock From amateurs to connoisseurs: modeling the evolution of user
  expertise through online reviews.
\newblock In \emph{Proceedings of the 22nd International Conference on World
  Wide Web}, pp.\  897--908, 2013.

\bibitem[M{\"u}ller et~al.(2019)M{\"u}ller, Kornblith, and
  Hinton]{muller2019does}
M{\"u}ller, R., Kornblith, S., and Hinton, G.~E.
\newblock When does label smoothing help?
\newblock In \emph{Proceedings of Advances in Neural Information Processing
  Systems}, pp.\  4696--4705, 2019.

\bibitem[Naeini \& Cooper(2016)Naeini and Cooper]{naeini2016binary}
Naeini, M.~P. and Cooper, G.~F.
\newblock Binary classifier calibration using an ensemble of near isotonic
  regression models.
\newblock In \emph{Proceedings of the 16th IEEE International Conference on
  Data Mining}, pp.\  360--369. IEEE, 2016.

\bibitem[Naeini et~al.(2015)Naeini, Cooper, and
  Hauskrecht]{naeini2015obtaining}
Naeini, M.~P., Cooper, G., and Hauskrecht, M.
\newblock Obtaining well calibrated probabilities using {B}ayesian binning.
\newblock In \emph{Proceedings of the 29th AAAI Conference on Artificial
  Intelligence}, 2015.

\bibitem[Niculescu-Mizil \& Caruana(2005)Niculescu-Mizil and
  Caruana]{niculescu2005predicting}
Niculescu-Mizil, A. and Caruana, R.
\newblock Predicting good probabilities with supervised learning.
\newblock In \emph{Proceedings of the 22nd International Conference on Machine
  Learning}, pp.\  625--632. ACM, 2005.

\bibitem[Nixon et~al.(2019)Nixon, Dusenberry, Zhang, Jerfel, and
  Tran]{nixon2019measuring}
Nixon, J., Dusenberry, M.~W., Zhang, L., Jerfel, G., and Tran, D.
\newblock Measuring calibration in deep learning.
\newblock In \emph{Proceedings of the 32nd IEEE Conference on Computer Vision
  and Pattern Recognition Workshops}, pp.\  38--41, 2019.

\bibitem[Paszke et~al.(2019)Paszke, Gross, Massa, Lerer, Bradbury, Chanan,
  Killeen, Lin, Gimelshein, Antiga, et~al.]{paszke2019pytorch}
Paszke, A., Gross, S., Massa, F., Lerer, A., Bradbury, J., Chanan, G., Killeen,
  T., Lin, Z., Gimelshein, N., Antiga, L., et~al.
\newblock Py{T}orch: An imperative style, high-performance deep learning
  library.
\newblock In \emph{Proceedings of Advances in Neural Information Processing
  Systems}, pp.\  8024--8035, 2019.

\bibitem[Pennington et~al.(2014)Pennington, Socher, and
  Manning]{pennington-etal-2014-glove}
Pennington, J., Socher, R., and Manning, C.
\newblock {G}lo{V}e: Global vectors for word representation.
\newblock In \emph{Proceedings of the Conference on Empirical Methods in
  Natural Language Processing}, pp.\  1532--1543. Association for Computational
  Linguistics, 2014.

\bibitem[Platt(1999)]{platt1999probabilistic}
Platt, J.
\newblock Probabilistic outputs for support vector machines and comparisons to
  regularized likelihood methods.
\newblock \emph{Advances in Large Margin Classifiers}, 10\penalty0
  (3):\penalty0 61--74, 1999.

\bibitem[Radford et~al.(2019)Radford, Wu, Child, Luan, Amodei, and
  Sutskever]{radford2019language}
Radford, A., Wu, J., Child, R., Luan, D., Amodei, D., and Sutskever, I.
\newblock Language models are unsupervised multitask learners.
\newblock \emph{OpenAI Blog}, 1\penalty0 (8), 2019.

\bibitem[Rajpurkar et~al.(2016)Rajpurkar, Zhang, Lopyrev, and
  Liang]{rajpurkar2016squad}
Rajpurkar, P., Zhang, J., Lopyrev, K., and Liang, P.
\newblock {SQ}u{AD}: 100,000+ questions for machine comprehension of text.
\newblock In \emph{Proceedings of the Conference on Empirical Methods in
  Natural Language Processing}, pp.\  2383--2392, 2016.

\bibitem[Schwarz et~al.(1978)]{schwarz1978estimating}
Schwarz, G. et~al.
\newblock Estimating the dimension of a model.
\newblock \emph{The Annals of Statistics}, 6\penalty0 (2):\penalty0 461--464,
  1978.

\bibitem[Srivastava et~al.(2014)Srivastava, Hinton, Krizhevsky, Sutskever, and
  Salakhutdinov]{srivastava2014dropout}
Srivastava, N., Hinton, G., Krizhevsky, A., Sutskever, I., and Salakhutdinov,
  R.
\newblock Dropout: a simple way to prevent neural networks from overfitting.
\newblock \emph{Journal of Machine Learning Research}, 15\penalty0
  (1):\penalty0 1929--1958, 2014.

\bibitem[Vaswani et~al.(2017)Vaswani, Shazeer, Parmar, Uszkoreit, Jones, Gomez,
  Kaiser, and Polosukhin]{vaswani2017attention}
Vaswani, A., Shazeer, N., Parmar, N., Uszkoreit, J., Jones, L., Gomez, A.~N.,
  Kaiser, {\L}., and Polosukhin, I.
\newblock Attention is all you need.
\newblock In \emph{Proceedings of Advances in Neural Information Processing
  Systems}, pp.\  5998--6008, 2017.

\bibitem[Wolf et~al.(2019)Wolf, Debut, Sanh, Chaumond, Delangue, Moi, Cistac,
  Rault, Louf, Funtowicz, and Brew]{Wolf2019HuggingFacesTS}
Wolf, T., Debut, L., Sanh, V., Chaumond, J., Delangue, C., Moi, A., Cistac, P.,
  Rault, T., Louf, R., Funtowicz, M., and Brew, J.
\newblock Huggingface's transformers: State-of-the-art natural language
  processing.
\newblock \emph{arXiv preprint arXiv:1910.03771}, 2019.

\bibitem[Zadrozny \& Elkan(2001)Zadrozny and Elkan]{zadrozny2001obtaining}
Zadrozny, B. and Elkan, C.
\newblock Obtaining calibrated probability estimates from decision trees and
  naive {B}ayesian classifiers.
\newblock In \emph{Proceedings of the 18th International Conference on Machine
  Learning}, volume~1, pp.\  609--616. Citeseer, 2001.

\bibitem[Zadrozny \& Elkan(2002)Zadrozny and Elkan]{zadrozny2002transforming}
Zadrozny, B. and Elkan, C.
\newblock Transforming classifier scores into accurate multiclass probability
  estimates.
\newblock In \emph{Proceedings of the 8th ACM SIGKDD International Conference
  on Knowledge Discovery and Data Mining}, pp.\  694--699. ACM, 2002.

\end{thebibliography}
\bibliographystyle{icml2019}

\end{document}